\documentclass{llncs}

\usepackage{graphicx}
\usepackage{longtable}

\usepackage {listings}
\usepackage{array,booktabs}
\newcolumntype{L}{@{}>{\kern\tabcolsep}l<{\kern\tabcolsep}}
\usepackage{colortbl}
\usepackage{xcolor}
\usepackage{hyperref}

\begin{document}

This is a pre-print of the following chapter: Arvind W. Kiwelekar,  Geetanjali S. Mahamunkar,  Laxman D. Netak,  Valmik B Nikam, {\em Deep Learning Techniques for Geospatial Data Analysis}, published in {\bf Machine Learning Paradigms}, edited by George A. TsihrintzisLakhmi C. Jain, 2020, publisher Springer, Cham reproduced with permission of publisher Springer, Cham. 

The final authenticated version is available online at: \href{https://doi.org/10.1007/978-3-030-49724-8_3}{https://doi.org/10.1007/978-3-030-49724-8\_3}.

{\bf Cite this chapter as:}

Kiwelekar A.W., Mahamunkar G.S., Netak L.D., Nikam V.B. (2020) Deep Learning Techniques for Geospatial Data Analysis. In: Tsihrintzis G., Jain L. (eds) Machine Learning Paradigms. Learning and Analytics in Intelligent Systems, vol 18. Springer, Cham.  \href{https://doi.org/10.1007/978-3-030-49724-8_3}{https://doi.org/10.1007/978-3-030-49724-8\_3}.

\mainmatter 
\title{Deep Learning Techniques for Geospatial Data Analysis}

\author{Arvind W. Kiwelekar \inst{1} \and Geetanjali S. Mahamunkar \inst{1} \and Laxman D. Netak \inst{1} \and Valmik B Nikam \inst{2}} 
\institute{Department of Computer  Engineering\\
Dr Babasaheb Ambedkar Technological University\\ 
Lonere, Raigad-402103, India \\
\and 
Department of Information and Technology \\
Veermata Jijabai Technical Institute\\ 
Mumbai-4000019, India\\
\{awk,ldnetak\}@dbatu.ac.in, gsmahamunkar@gmail.com, vbnikam@it.vjti.ac.in}

\maketitle

 \begin{abstract}
 Consumer electronic devices such as mobile handsets, goods tagged with RFID labels, location and
position sensors are continuously generating a vast amount of location enriched data called
geospatial data. Conventionally such geospatial data is used for military applications. In recent
times, many useful civilian applications have been designed and deployed around such geospatial
data. For example, a recommendation system to suggest restaurants or places of attraction to a
tourist visiting a particular locality. At the same time, civic bodies are harnessing geospatial data
generated through remote sensing devices to provide better services to citizens such as traffic
monitoring, pothole identification, and weather reporting. Typically such applications are leveraged
upon non-hierarchical machine learning techniques such as Naive-Bayes Classifiers, Support Vector
Machines, and decision trees. Recent advances in the field of deep-learning showed that Neural
Network-based techniques outperform conventional techniques and provide effective solutions for
many geospatial data analysis tasks such as object recognition, image classification, and scene
understanding. The chapter presents a survey on the current state of the applications of deep learning techniques for analyzing geospatial data.

The chapter is organized as below: (i) A brief overview of deep learning algorithms. (ii)Geospatial Analysis: a Data Science Perspective (iii) Deep-learning techniques for Remote Sensing  data analytics
tasks (iv) Deep-learning techniques for GPS data analytics(iv) Deep-learning techniques for RFID data analytics.
 
 \end{abstract}

\section{Introduction}

Deep learning has emerged as a preferred technique to build intelligent products and services in various application domains.  The resurgence of deep learning in recent times is attributed to three key factors. The first one is the availability high-performance GPUs necessary to execute computation-intensive deep learning algorithms. Second, the availability of such hardware at an affordable price. Also, the third and most important key factor responsible for the success of deep learning algorithms is the availability of open datasets in various application domains such as ImageNet\cite{krizhevsky2012imagenet} required to train the deep learning algorithms extensively\cite{goodfellow2016deep}.

The conventional application domains in which deep learning techniques have been applied effectively are speech recognition, image processing, language modelling and understanding,  natural language processing and information retrieval \cite{deng2014tutorial}. All these application domains include processing and retrieving of useful information from raw multimedia data.

The success of deep learning techniques in these fields triggered its application in other fields such as Biomedicine \cite{cao2018deep},  Drug Discovery \cite{chen2018rise},  and Geographical  Information System \cite{zhang2016deep}.

The chapter presents a state of the art review on applications of deep learning techniques for geospatial data analysis, one of the fields which is increasingly applying deep learning techniques to understand our planet earth.

\section{Deep Learning: A Brief Overview}
The field of Deep Learning is a sub-field of  Machine Learning which studies the techniques for establishing a relationship between input feature variables and one or more output variables.  Many tasks, such as classification and prediction, can be represented as a mapping between input feature variables and output variable(s). 

For example, the price of a house in a city is a function of input feature variables such as the number of rooms,  built-up area, the locality in a city, and other such parameters. The goal of the machine learning algorithms is to learn a mapping function from the given input data set, which is referred to as a training data set.

For example, the variables  $x_{1}$, $x_{2}$,.... $x_{n}$ which are input feature variables and  the variable  $y$  which is an output variable can be represented as 
 
\[y = f(x_{1}, x_{2},.... x_{n}) \]

The input variables are also referred to as independent variables, features, and predictors. The function $f$ is referred to as a {\em model} or {\em hypothesis function}.  

The training data set may or may not include the values of $y$, i.e., output variable. When a machine learning algorithm learns the function $f$ from both features and output variable, the algorithm is referred to as a {\em supervised} algorithm. An {\em unsupervised learning} algorithm learns the function $f$ from input features only without knowing the values of the output
variable. Both kinds of the algorithm have been widely in use to develop intelligent product and services.

There exist many machine learning algorithms (e.g., Linear Regression, Logistic Regression, Support Vector Machine \cite{Domingos:2012:FUT:2347736.2347755}) which are practical to learn simple tasks such as predicting house prices based on input features(e.g. the number of bedrooms, built-up area, locality).  The goal of these learning algorithms is to reduce the error in predicting the value of the output variable. This minimizing goal is captured by a function called  {\em cost function}. The
{\em stochastic gradient descent} is one of the optimization techniques that is commonly used to achieve
the minimization goal. 

The conventional machine learning algorithms have been found useful and effective in learning simple tasks such as predicting house prices and classifying the tumour as malignant or benign.  
In such situations,  the relationship between input features and the output variable is a simple linear function involving few predefined features.

However, they fail to perform effectively in situations where it is difficult to identify the features required for prediction — for example, recognizing the plant, rivers, animals in a given image.  In such situations, the number of input features required for accurate prediction is large,  and the relationship between input features and the output variable is also complex and non-linear one.

The set of algorithms that belongs to Deep Learning outperforms as compared to conventional machine learning methods in such situations.  This section briefly  describes deep learning techniques that have been found useful for geospatial data analysis. For more detailed and elaborate discussion on these techniques, one can refer \cite{goodfellow2016deep}.

\begin{figure}[t]

\center{\includegraphics[scale=0.35]{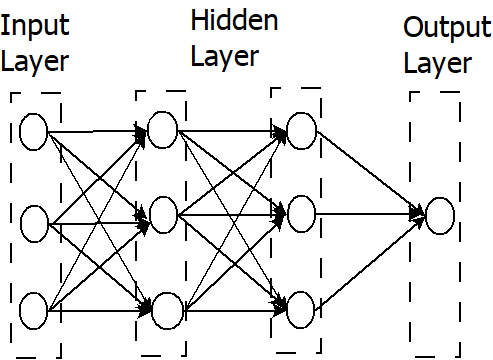}}
  \caption{Deep Neural Network.}
  \label{ann}
\end{figure}

\subsection{Deep Learning Architectures}

The deep learning techniques attempt to reproduce or emulate the working of the human brain in an artificial context.  A network of neurons is the structural and functional unit of the human brain.  Likewise, the Artificial Neural Network (ANN) is the fundamental element underlying most of the deep learning techniques. 

\subsection{Deep Neural Networks}
A simple ANN consists of three essential elements: (i) Input Layer (ii) Hidden Layer and (iii) Output Layer. An input layer consists of values of input features, and an output layer consists of values of output variables.  A hidden layer is referred to as hidden because values held by the neurons in a hidden layer are not visible during information processing.

 A layer consists of more than one information processing nodes called {\em neurons}. An artificial neuron is an information processing node taking $n$-inputs and producing $k$-outputs. It is essentially a mapping function whose output is defined in two steps. The first step is the sum of weighted multiplication of all of its input. Mathematically, it can be represented as:
\[Y_{i} =g( \sum_{j}W_{ij}*a_{j}),\] 
where $Y_{i}$ is output of $i^{th}$ node, $W_{ij}$ is the weight of $j^{th}$ input on  $i^{th}$ node and $a_{j}$ is the value of  $j^{th}$ input.  This operation implements a matrix multiplication operation which is a linear function. 
In the second step, the output of the first step is fed to a non-linear function called an {\em activation function}. A neural network may use any one of the functions (i) Sigmoid function (ii) Hyperbolic tangent function or (iii) Rectified Linear Unit (ReLU)

The modern {\em deep neural networks} consist of more than one hidden layer, as shown in Figure \ref{ann}  and  ReLU as an activation function. 

Training of the DNN is an iterative process which usually implements a stochastic gradient algorithm to find the parameters or weights of input features.  The weights of the parameters are randomly initialized or used from a  pre-trained model. An error in the prediction is calculated at the output layer. At a hidden layer, the gradient of error which is a partial derivative of the error with respect to the existing values of weights is fed back to update the values of weights at the input layer during next iteration. The process is known as back-propagation of gradients.

This seemingly simple strategy of learning parameters or weights of a model works effectively to detect features required for classification and prediction in many image processing and speech recognition tasks, provided that hardware required to do matrix multiplication and a large data-set is available. Thus eliminating the need for manual feature engineering required for non-hierarchical classification and prediction mechanism. 

The deep neural networks have been successfully used to detect objects such as handwritten digits, and pedestrians \cite{lecun2015deep}. In general, DNNs have been found efficient in handling 2-dimensional data that can be represented through a simple matrix.

\begin{figure}[t]
\center{\includegraphics[scale=0.35]{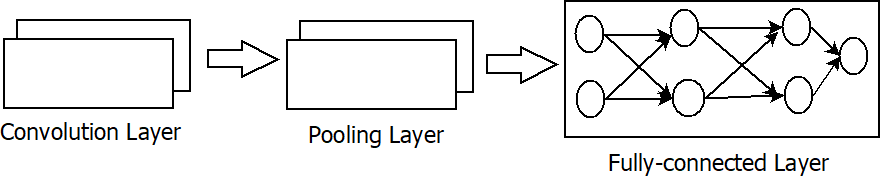}}
  \caption{Convolutional Neural Network.}
  \label{cnn}
\end{figure}

\subsection{Convolutional Neural Network (CNN)}
As seen in the previous section, the input of a hidden layer of DNN  is connected to all the outputs of the previous layer making computations and handling the number of connections unmanageable in case of high-dimensional matrices. Such situations arise when image sizes are of high resolutions (1024 X 1024). Also, the DNNs perform well when the data-set is 2-dimensional.  But the majority of multi-modal data sets, for example, coloured images, videos, speech, and text are of 3-dimensional nature. 

The {\em Convolutional Neural Networks} (CNN) are employed when the data set is three dimensional in nature and matrix size is very large. For example, a high-resolution coloured image has three channels of pixels (i.e., Red, Blue, Green) of size 1024 X 1024.

The architecture of CNN, as shown in Figure \ref{cnn} can be divided into multiple stages.  These stages perform pre-processing activities such as identifying low- level features and reducing the number of features,  followed by the main task of classification and prediction as done by a fully connected neural network.

The pre-processing activities are done by  {\em convolution layers} and  {\em pooling layers}. 

The {\em convolution layer} performs step-wise convolution operation on the given input data size and a filter bank to create a  feature map.  Here, a filter is a matrix of learn-able weights representing a pattern or a motif.  The purpose of the step is to identify low-level features — for example,  edges in an image.  The underlying assumption of this layer is that low-level features correlate with a pixel configuration \cite{lecun2015deep}.

The {\em pooling layer} implements an aggregation operation such as either addition or maximization or average with an intention to share the weights among multiple connections. Thus reducing the number of features required for classification and/or prediction.  

These pre-processing stages drastically reduce the number of features in a fully connected part of CNN.

The CNNs have been found useful in many image processing and speech recognition tasks. Because the working of CNN is based on the assumption that many high-level features are the compositions of low-level features. For example, a group of edges constitute a pattern or a motif, a group of motif constitute a part of an image,  and a group of parts constitute an object \cite{lecun2015deep}.

\begin{figure}[t]

\center{\includegraphics[scale=0.3]{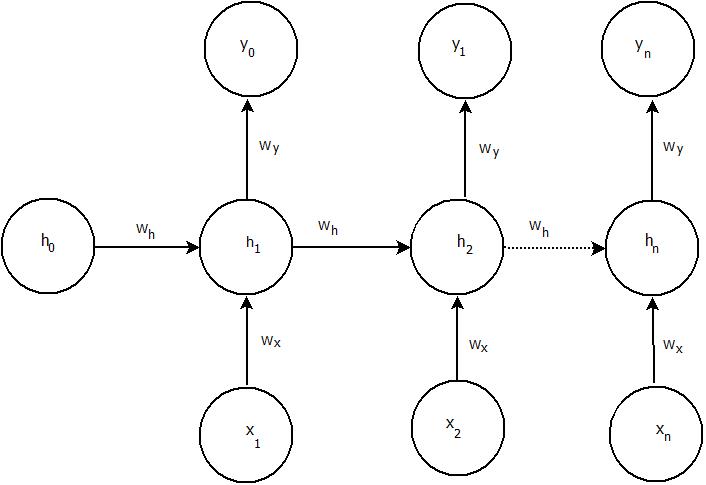}}
  \caption{Recurrent Neural Network.}
  \label{rnn}
\end{figure}

\subsection{Recurrent Neural Networks (RNN)}
The third type of neural network is Recurrent Neural Networks (RNN). It has found applications in many fields such as speech  recognition and machine language translation.

The RNN learns the dependency relationship among the sequence of input data. Unlike DNN and CNN, which learn relationships among feature variables and output variables, the RNN learns the relationship between data items fed to the network at different instances.  To do this, RNN maintains a state vector or memory, which is implemented by connecting the hidden layer of current input to the hidden layer of previous input, as shown in Figure \ref{rnn}.  Hence, the output of RNN is not only the function of current input data but also the input data observed so far. As a result, the RNN may give two different outputs for the same input at two separate instances.

A variant of RNN called RNN with LSTM (Long Short Term Memory) uses a memory unit to remember long term dependencies between data items in a sequence.  Such networks have been found useful in question answering systems and symbolic reasoning, which draw a conclusion from a series of premises.

\begin{figure}[t]

\center{\includegraphics[scale=0.3]{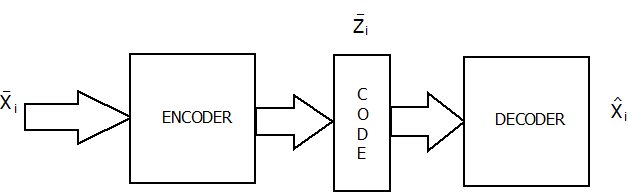}}
  \caption{Auto-Encoders}
  \label{ae}
\end{figure}

\subsection{Auto-Encoders  (AE)}
The autoencoder is an {\em unsupervised} deep neural network architecture.  Unlike supervised deep neural networks (e.g., DNN, CNN, RNN) which establish a mapping between input and output variables, autoencoders identify patterns in input data intending to transform the input data items.  Reducing the dimensions of input vector is one example of transformation activity. In such variants, autoencoders transform an n-dimensional input vector into k-dimensional output vector such that  $k<n$.  
These networks learn two different hypothesis function called {\em encode} and {\em decode}.  The purpose of the function {\em encode} is to transform input vector $x$ in some other representation $z$ which is  defined as
\[ z= encode(x)  \] where  $z$, and $x$ are vectors with different representations.
Similarly, the purpose of the function $decode$ is to restore the vector $z$ to its original form:
 \[ x= decode(z) \]
 
 The functions $encode$ and $decode$ may be simple linear functions or a complex non-linear function. 
 In case the data is highly non-linear an architecture called deep autoencoder with hidden layers is employed.
 
 Autoencoders are typically employed for dimensionality reduction and pre-training a deep neural network. It has been observed that the performance of CNN or RNN improves when they are pre-trained with autoencoders \cite{le2015tutorial}.
 
\section{Geospatial Analysis:A Data Science Perspective}
The purpose of geospatial data analysis is to understand the planet Earth by collecting, interpreting and visualizing the data about objects, properties and events happening on, above and below the surface of Earth. Tools that we use to manage this information influence our knowledge of the Earth. So this section briefly reviews the technologies that enable geospatial data analysis.
\subsection{Enabling Technologies for Geospatial Data Collection}

The tools that are being used to collect location-specific information  include (i) Remote Sensing (ii) Unmanned Aerial Vehicles (UAV) (iii) Global Positioning System (GPS) and (iv) Radio Frequency Identifiers (RFID). This section briefly explains these techniques.

\begin{enumerate}
\item {\em Remote Sensing}

Remote sensing is a widely used technique to observe and acquire information about Earth without having any physical contacts. The principle of electromagnetic radiation is used to detect and observe objects on Earth.   With Sun as the primary source of electromagnetic radiations, sensors on remote sensing satellites detect and record energy level of radiations reflected from the objects on Earth. 

Remote sensing can be either active or passive.  In {\em passive} remote sensing, the reflected energy level is used to detect objects on Earth while in {\em active} remote sensing the time delay between emission and delay is used to detect the location, speed, and direction of the object.

The images collected by a remote sensor are characterized primarily by attributes such as {\em spectral resolution} and {\em spatial resolution}.  Spatial resolution describes the amount of physical area on Earth corresponding to a pixel in a raster image. Typically it corresponds to  1 to 1000 meter area. Spectral resolution corresponds to the number of electromagnetic bands used in a pixel which typically corresponds to 3  bands (e.g., Red, Green and Blue ) to seven visible colours. In hyper-spectral imaging produced by some  remote sensing mechanism, 100 to 1000 bands correspond to a pixel.

Data acquired through remote sensing have been useful in many geospatial analysis activities such as precision agriculture\cite{seelan2003remote}, in hydrology \cite{jackson1996remote}, for monitoring soil moisture \cite{wang2009satellite} to name  a few.

\item{\em Drones and Unmanned Aerial Vehicles (UAV)}

Drones and UAVs are emerging as a cost effective alternative to conventional satellite-based method of remote sensing for Earth surface imaging \cite{themistocleous2014use}. Like satellite based remote sensing, UAVs are equipped with multi-spectral cameras, infrared cameras, and thermal cameras. A spatial accuracy of 0.5m to 2.5 meter has been reported when UAVs are used for remote sensing \cite{rokhmana2015potential}. Remote sensing with UAVs have found applications for precision agriculture \cite{rokhmana2015potential} and civilian security applications \cite{daniel2011using}.

\item {\em Global Positioning Systems (GPS)}

The devices equipped with Global Positioning Systems (GPS) can receive signals from GPS satellites and can calculate its accurate position. Smartphones, Cars and dedicated handheld devices are examples of GPS.  These devices are specifically used for navigation purpose showing directions to a destination on maps, and monitoring and tracking of movements of objects of interest. Such devices can identify their locations with an accuracy of 5 meters to 30 centimeters.  
The location data collected from GPS devices have been used to identify driving styles \cite{dong2016characterizing}, to predict traffic conditions \cite{niu2014deepsense} and transportation management.
\item  {\em Radio Frequency Identification} (RFID) 

It is a low-cost technological alternative to GPS used for asset management. It is typically preferred when assets to be managed move around a shorter range. Unlike GPS device, RFID tags are transmitter of radio waves which are received by RFID tracker to identify the location of RFID tagged device. Recently  data generated by RFID devices have been found useful to recognise human activities\cite{li2016deep}, and to predict order completion time \cite{wang2019deep}.

\end{enumerate}

\subsection{Geospatial Data Models}
The geospatial data models represent information about earth surface and locations of objects of reference. The data model used to represent this information  depends on the mode used to collect the data. The  raster and vector data models are used, when the mode of data collection used  is remote sensing and  UAV, while, GPS and RFID data models are used when mode of data collection GPS and RFID respectively.

The three data models that are prevalent in the field of remote sensing and UAV are raster, vector and Triangular Irregular Networks (TIN). In raster data model the earth surface is represented through points, lines and polygons.  In vector representation, the earth surface is represented through cell matrices that store numeric values. In TIN  data model the earth surface is represented as non-overlapping contiguous triangles.

The GPS data contains the location information of GPS enabled device in terms of longitude, latitude, timestamps and the satellites used to locate the device.

The RFID data contains information about Identification number of the RFID tag, location and timestamp.
\subsection{Geospatial Data Management}
Geographic Information System (GIS) integrates various aspects of geospatial analysis into one unified tool. It combines multiple geospatial analysis activities such as capturing of data, storage of data, querying data, presenting and visualizing data. It provides a user interface for users to perform these activities.

GIS integrates various data capturing technologies such as satellite-based remote sensing, UAV-based remote sensing, GPS devices,  and scanning of paper-based maps.

GIS uses multiple kinds of data models such as raster data model, vector data model to represent earth surfaces as a base map. On top of a base map layers are used to prepare thematic maps showing roads, land cover and population 
distributions.
Few examples of GIS are ArcGIS \cite{agis}, Google Earth \cite{ge}, QGIS\cite{qgis}.

\section{Deep learning for Remotely Sensed Data Analytics}
Images constitute the significant chunk of data acquired through the method of remote sensing.   These images vary in terms of information representation and graphical resolution.  Typically the images obtained through remote sensing represent information either in vector or in raster form.   Depending on the resolution of the images, they can be of Low Resolution (LR), High Resolution(HR) and Very High Resolution(VHR) type. 

Deep learning techniques have been found as a promising method to process images of all kinds.  Especially for the purpose of image segmentation\cite{segment}, image enhancement \cite{enhance},  and image classification\cite{classification},  and to identify objects in images \cite{object}.

Applications of deep learning techniques  for remotely sensed data analytics leverage these advancements in the field of image processing to develop novel applications for geospatial data analysis.  This section briefly reviews some of these recent applications. A detailed survey appears in \cite{ma2019deep,zhu2017deep}.

\subsection{Data Pre-processing} The images obtained through remote sensing techniques often are of poor quality due to atmospheric conditions. The quality of these images needs to be enhanced to extract useful information from these images. 

A set of activities which include denoising, deblurring, super-resolution, pan-sharpening and image fusion are typically performed on these images. Recently, geoscientists have started applying deep learning techniques for this purpose. Table \ref{tab1} shows application of deep learning techniques for image pre-processing activities.

\begin{table}[t]
    \centering
    \begin{tabular}{|p{1in}|p{1.5in}|p{1.5in}|}
       \hline 
       Pre-Processing Activity & Example & Deep learning  Technique used  \\ \hline 
         Denoising &  To Restore original clean image from the low quality image or image with irrelevant details. & A combination of sparse coding with denoising Auto Encoders \cite{xie2012image} denoising CNN\cite{zhang2017beyond}. \\ \hline 
         
         Deblurring & Restoring original or sharp image from the blurred image. The blurring of images occurs due to atmospheric turbulence in remote sensing. & CNN\cite{nah2017deep}, A combination of Auto-Encoders and Generative Artificial Neural Network (GAN) \cite{nimisha2017blur}. \\ \hline 
         Pan Sharpening, Image Fusion, Super Resolution &  In many geospatial applications images with high spatial and high spectral resolutions are required. The {\em pan sharpening} is a method to combine a LR  multi-spectral image with  a HR panchromatic image. The {\em super-resolution} combines  a LR hyper-spectral (HS) image and a HR MS image. &  Stacked Auto-Encoders \cite{liu2018deep}, CNN \cite{masi2016pansharpening}. \\ \hline 
    \end{tabular}
    \caption{Application of Deep learning Image Pre-processing}
    \label{tab1}
\end{table}
From the table, it can be observed that both supervised(e.g., CNN) and  unsupervised  (e.g., Auto-encoders) are used for image pre-processing.  Automatic feature extraction and comparable performance with conventional methods are some of the motivating factors behind adopting DL techniques for image pre-processing. 

\subsection{Feature Engineering}
In the context of deep learning, feature engineering  is simple as compared with conventional machine learning techniques because many deep learning techniques such as CNN automatically extract  features. This is one reason to prefer deep learning techniques for the analysis of remote sensing imagery.  However, there exists other feature engineering steps described below which need to be carried out for better performance and reuse of knowledge learnt on similar applications.
\begin{enumerate}
  \item {\em Feature Selection}: Feature selection is one of the crucial steps in the feature engineering aimed to identify the most relevant features that contribute to establishing the relationship between input and output variables.  Earlier studies observe that three key factors affect the performance of machine learning techniques\cite{yan2019structured}. These are  (i)choice of data set, (ii) machine learning algorithm and (iii) features used for classification or prediction tasks.    In conventional non-hierarchical machine learning methods, regression analysis is usually performed to select the most relevant features.
  
  Recently, a two stage DL-based technique to select features is proposed in \cite{dong2019revisiting}. It is  based on a combination of  supervised deep networks and autoencoders. Deep networks in the first stage  learn complicated low-level representations. In the second stage,  an unsupervised autoencoders  learn simplified representations. Few special DL-techniques specific to geospatial application have also been developed in order to  extract spatial and spectral features\cite{zhao2016spectral,zou2015deep} from remotely sensed imagery  
  
  \item {\em Feature Fusion:}
  Feature fusion is the process of identifying {\em discriminating features} so that an optimal set of features can be used to either classify or predict an output variable.   For example, feature fusion is performed to recover a high-resolution image from a low-resolution image \cite{wang2019cfsnet}. 
A remote sensing specific DL- technique is proposed in \cite{li2019multiscale}.  The method is designed to classify hyper-spectral images.  In the first stage, an optimal set of multi-scale features for CNN are extracted from the input data. In the second stage, a collection of discriminative features are identified by fusing the multi-scale features.
  
    \item {\em Transfer learning or domain Adaptation}: The problem of domain adaptation is formulated with respect to   the classification of remotely sensed images in \cite{tuia2016domain}.  It is defined as adapting a   classification model to spectral properties or spatial regions that are different from the images used for training purposes. Usually, in such circumstances, a trained model fails to accurately classify images because of different acquisition method or atmospheric conditions, and they are called to be sensitive to data shifts. Domain adaptation is a specialized technique to address the transfer learning problem.  Transfer learning deals a general situation where the features of a trained classifier model are adapted to classify or predict data that is not stationary over a period of time or space. For example, images captured through a remote sensing device vary at day and night time.
    
    A strategy  called  {\em transduction strategy}  that adopts  DL techniques (e.g., autoencoders) has been proposed in \cite{bengio2012deep} to address the problem of transfer learning.  The strategy suggests to pre-train a DL architecture on a test data set using unsupervised algorithms (e.g., autoencoders) to identify discrimating features.  Here, the test data set is used to pre-train a DL architecture and the test data set  is entirely different from the training data set used to train a DL architecture. The strategy works to learn abstract representations because it  identifies a set of  discriminating features responsible to generalize the inferences from the observed the data.
\end{enumerate}
\subsection{Geospatial Object Detection}
The  task of detecting or identifying an object of interest such as a road, or an airplane, or a building from aerial images acquired either through remote sensing or UAV is referred to as geospatial object detection.

It is a special case of general problem of object recognition from images. However with numerous challenges such as small size of the object to be detected, large number of objects in imagery, and complex environment \cite{cira2019deep,zhang2016deep} make the task of object recognition more difficult. Hence,  DL-methods have been found useful in such situations specially  to extract low-level features and learn abstract representations.

When DL-based methods are applied, the models such as CNN  are first trained with geospatial images labeled to separate out the objects of our interest. The images used for training purposes are optimized to detect a specific object. These steps can be repeated until the image analysis model accurately detect multiple types of objects from the given images \cite{estrada2018broad}.

\subsection{Classification Tasks in Geospatial Analysis}

Like in other fields such as Computer Vision and Image processing, the field of Geospatial Analysis primarily adopts DL-techniques to classify remotely sensed imagery in various ways to extract useful information and patterns.  Some of these usecases are discussed below.

These classification tasks are either {\em pixel-based} or {\em object-based} \cite{weihobject}. In pixel-based classification, pixels are grouped based on spectral properties (e.g., resolution). In object based classification, a group pixels are classified using both spectral and spatial properties into various geometric shapes and patterns.  Either a supervised (e.g., CNN) or unsupervised algorithms (e.g. Autoencoders)   are applied for   the purpose of classification.  In comparison with conventional machine learning techniques, application of DL-based techniques has outperformed in both scenarios in terms of classification accuracy\cite{zhang2016deep}. 

Some of the classification activities include. 
\begin{enumerate}
    \item {\em Land Cover Classification}:
    
    Observed bio-physical cover over the surface of earth is referred to as {\em land cover}. A set of 17 different categories have been identified by  International Geosphere-Biosphere Programme (IGBP) to classify earth surface. Some of these broad categories include Water, Forest, Shrubland, Savannas, Grassland, Wetland, Cropland, Urban, Snow, and Barren \cite{di2005land}.
    
     There are two primary methods of getting information about land cover: field survey and remotely sensed imagery. DL-techniques are applied when information is acquired through remote sensing. In \cite{xu2017automatic,kussul2017deep,scott2017training} the pre-trained CNN model  is used  to classify different land cover types.  CNN is the most preferred technique to classify earth surface according to land cover. However other techniques such as RNN are also applied in few cases\cite{sun2019using}.
     \item {\em Land Use Classification}:
     
     In this mode of classification, earth surface can be classified according to their social, economical and political usages. For example, a piece of green-land may be used as a tennis court or a cricket pitch. A piece of urban can be used for industrial or residential purposes. So there are no fixed set of categories or a classification system to differentiate land uses.
     
     Like in case of land cover classification, CNN \cite{helber2019eurosat} is typically used to classify land uses and to detect change in land cover according their uses \cite{cao2019land}. 
     \item {\em Scene Classification}:
     
     Scenes in remotely sensed images are high-level semantic entities such as a densely populated area,  a river, a freeway, a golf-course, an airport etc. Supervised and unsupervised deep learning techniques have been found effective for scene understanding and scene classification in remotely sensed imagery \cite{cheng2017remote}. 
\end{enumerate}
Further, DL-techniques have been applied in specific cases of land use and land cover classification such as fish species classification \cite{salman2016fish}, crop type classification \cite{kussul2017deep} and mangrove classification \cite{faza2018}.

\section{Deep learning for GPS Data Analytics}

GPS enabled devices, particularly vehicles, are generating data in the range of GB/Sec. Such data includes information about exact coordinates, i.e. longitude and latitude of the device, direction and speed with which the device is moving. Such information throws light on many behavioural patterns useful for targetted marketing.

One way to utilize this data is to identify the driving style of a human driver.  Various methods of accelerating, braking, turning under different road and environmental conditions determine the driving style of a driver.  This information is useful for multiple business purposes like insurance companies may use to mitigate driving risks, to design better vehicles, to train autonomous cars,  and to identify an anonymous driver.

 However, identifying features that precisely determine a driving style is a challenging task.  Hence the techniques based on deep learning are useful to extract such features. W. Dong et al. \cite{dong2016characterizing} have applied DL-techniques to GPS data to characterize the driving styles.  The technique is motivated by the applications of DL-methods in speech recognition.  They interpreted GPS data as a time series and applied CNN and RNN to identify driving styles.  The raw data collected from the GPS devices are transformed to extract statistical features (e.g., mean and standard deviation) and then it is fed for analysis to CNN and RNN. The method was validated through identifying the driver's identity.

In another interesting application, a combination of RNN and Restricted Boltzman machine \cite{ma2015large} is adopted to analyze GPS data for predicting the congestion evolution in a transportation network.  The technique effectively predicts how congestion at a place has a ripple effect at other sites.

GPS data has also been analyzed using DL techniques to manage resources efficiently. For example, in the construction industry, the usages of GPS enabled equipment are monitored and analyzed \cite{pradhananga2013automatic}.

\section{Deep learning for RFID Data Analytics}

 RFID and GPS are emerging as technologies to ascertain the locations of devices or assets. However, RFID in comparison with GPS allows to build location-sensitive applications for shorter geographical coverage.  Unlike GPS mechanism,  RFID is a low cost alternative to measure locations of assets with respect to a RFID tracker. GPS enabled devices  provide  absolute location information in terms of longitude and latitude.
 
 The RFID's potential to accurately measure the location of devices with respect to a  RFID tracker is increasingly being used in automated manufacturing plants. By using the location data transmitted by a RFID tagged device, various intelligent applications have been designed   by applying DL techniques. These applications are intended to identify patterns and recognise activities in a manufacturing or a business process.  
 
 DL techniques have been found specially effective when activities happens in a spatio-temporal dimension.  For example,  in the field of medicine, detecting activities during trauma resuscitation \cite{li2016deep}.
 
 In such situations,  activity recognition is represented  as a multi-class classification problem. For example, in case of trauma resuscitation,  activities are oxygen preparation, blood pressure measurement, temperature measurement, and cardiac lead placement. For detecting these activities, a hardware set up of RFID tagged device along-with RFID tracker is used to collect the data. The collected data is analysed using CNN to extract relevant features and recognise the activity during trauma resuscitation.
 
 In another  application from manufacturing processes, the deep learning techniques have been used to accurately predict the job completion time. The conventional methods of job completion time rely on use of historical data. Such predictions greatly vary from the actual job completion time.  The method proposed in \cite{wang2019deep} adopts RFID tags and trackers to collect real-time data. The collected data then mapped to historical data using CNN for predicting job completion time.

\section{Conclusion}

The scope of the geospatial data analysis is very vast. The data collection methods vary from manual one to satellite-based remote sensing. Numerous data models have been developed to represent various aspects of earth surfaces and objects on earth. These models capture absolute and relative location-specific information. These models also reveal static and dynamic aspects, spectral and spatial resolutions.  Various GIS tools are being used to integrate and manage geospatial information.

So far, harnessing this collected information for useful purposes such as to extract hidden information in the form patterns, behaviours and predictions were limited by the absence of powerful analysis methods. But the emergence of Deep learning-based data analysis techniques has opened up new areas for geospatial applications.

This chapter presents some of the emerging applications designed around DL-techniques in the field of geospatial data analysis. These applications are categorized based on the mode of data collection adopted.

 The deep-learning architectures such as CNN and Autoencoders are increasingly used when the method of data collection is remote sensing and UAV.   Applications of DL techniques realize the high-level classification tasks such as land uses and land covers. 

When GPS is the primary method of data collection, the collected data needs to be interpreted as a sequence or as a series of data. In such situations, RNN, along with CNN, is increasingly used to identify hidden patterns and behaviours from the traffic and mobility data collected from GPS enabled devices.

The CNNs are primarily used to process the location-specific information gathered through RFID device over shorter geographic areas. These analyses lead to recognize a Spatio-temporal activity in a manufacturing or a business process and to predict the time required to complete these activities using real-time data, unlike  using historical data for predictive analytics.

These novel applications of DL-techniques in the field of Geospatial Analysis are improving our knowledge of planet earth, creating new insights about our behaviour during mobility, assisting us to make decisions (e.g.,  route selection), and making us more environmentally conscious citizens (e.g., predictions about depleting levels of forest land and protections of mangroves). 

\subsection*{Acknowledgement:} The authors acknowledge the funding provided by Ministry of Human Resource Development (MHRD), Government of India, under the Pandit Madan Mohan National Mission on Teachers Training (PMMMNMTT). The work presented in this chapter  is based on the course material developed to train engineering teachers  on the topics of Geospatial Analysis and Product Design Engineering. 
\bibliographystyle{plain}
\bibliography{gsda}

\end{document}